\documentclass[sigconf]{acmart}
\usepackage{etex}
\renewcommand\footnotetextcopyrightpermission[1]{\emph{}} 
\usepackage{todonotes}
\usepackage{booktabs} 
\usepackage[plain]{algorithm}
\usepackage[noend]{algpseudocode}
\usepackage{microtype}
\usepackage{pgfplots}
\usepackage{pstricks-add}
\usepackage{multirow}

\begin{document}
\title{Streamlined Deployment for Quantized Neural Networks}

\author{Yaman Umuroglu, Magnus Jahre}
\affiliation{%
  \institution{Norwegian University of Science and Technology}
  \city{Trondheim, Norway}
}
\email{yamanu@ntnu.no}

\renewcommand{\shortauthors}{Y. Umuroglu et al.}

\newcommand{\ours}{\textsc{BitSerialGEMM}}
\newcommand{\qnn}[2] {$\mathbf{W^#1A^#2}$}
\newcommand{\bitplane}[2] {$#1[#2]$}
\newcommand{\pluseq}{\mathrel{+}=}
\newcommand{\myceil}[1]{\left \lceil #1 \right \rceil }
\newcommand{\TODO}[1]{\todo[inline]{#1}}

\begin{abstract}
Running Deep Neural Network (DNN) models on devices with limited computational capability is a challenge due to large compute and memory requirements.
Quantized Neural Networks (QNNs) have emerged as a potential solution to this problem, promising to offer most of the DNN accuracy benefits with much lower computational cost.
However, harvesting these benefits on existing mobile CPUs is a challenge since operations on highly quantized datatypes are not natively supported in most instruction set architectures (ISAs).
In this work, we first describe a \emph{streamlining} flow to convert all QNN inference operations to integer ones.
Afterwards, we provide techniques based on processing one bit position at a time (bit-serial) to show how QNNs can be efficiently deployed using common bitwise operations.
We demonstrate the potential of QNNs on mobile CPUs with microbenchmarks and on a quantized AlexNet, which is $3.5\times$ faster than an optimized 8-bit baseline.
Our bit-serial matrix multiplication library is available on GitHub.
\end{abstract}

\maketitle
\setlength{\textfloatsep}{5pt}

\section{Introduction}

From voice recognition to object detection, \emph{Deep Neural Networks (DNNs)} are steadily getting better at extracting information from complex raw data.
Combined with the popularity of mobile computing and the rise of the Internet-of-Things (IoT), there is enormous potential for widespread deployment of intelligent devices, but a computational challenge remains.
A modern DNN can require billions of floating point operations to classify a single image, which is far too costly for energy-constrained mobile devices.
Offloading DNNs to powerful servers in the cloud is only a limited solution, as it requires significant energy for data transfer and cannot address applications with real-time or low-latency requirements, such as augmented reality or navigation for autonomous drones.

\emph{Quantized Neural Networks (QNNs)} have recently emerged as a potential solution to this problem.
They contain convolutional, fully-connected, pooling and normalization layers similar to the floating point variants, but use a constrained set of values to represent each weight and activation in the network.
We will use the notation \qnn{w}{a} to refer to a QNN with $w$-bit weights and $a$-bit activations, and focus on cases where they represent \emph{few-bit integers} ($w, a \leq 4$).
The computational advantages of such QNNs are two-fold:
\begin{enumerate}
	\item Each parameter and activation can be represented with a few bits.
	A greater portion of the working set can thus be kept in on-chip memory, enabling greater performance, reducing off-chip memory accesses and the energy cost of data movement.
	\item Most QNN operations are on few-bit integers, which are faster and more energy-efficient than floating-point.
\end{enumerate}

While a quantized network will generally have reduced accuracy compared to an equivalent DNN using floating point, recent research has demonstrated significant progress in closing this accuracy gap.
Courbariaux and Hubara et al. \cite{binarynet} first demonstrated that Binarized Neural Networks (BNNs), a QNN variant with \qnn{1}{1}, could achieve competitive accuracy on smaller image recognition benchmarks like CIFAR-10 and SVHN.
XNOR-Net \cite{xnornet} improved upon this technique by adding scaling factors to better approximate the full-precision operations.
Noting that more challenging classification tasks such as ImageNet could benefit from higher-precision activations, DoReFa-Net \cite{dorefanet} used multi-bit activations and weights to further improve accuracy.
Recently, Cai et al. \cite{hwgq} proposed Half-wave Gaussian Quantization (HWGQ) to take advantage of the Gaussian-like distribution of batch-normalized activations, demonstrating \qnn{1}{2} networks with less than 5\% top-5 accuracy drop compared to floating point DNNs on the challenging ImageNet dataset, as summarized in Table \ref{tab:hwgq-accuracy}.

Despite the attractive accuracy and computational properties, there is a challenge in reaping the benefits on QNNs on mobile devices with commodity processors.
Three outstanding issues limit the benefits of QNN deployment on existing mobile CPUs: floating point parameters inside and between quantized layers, lack of native support for efficient few-bit integer matrix multiplications, and overhead of bit-masking operations for convolution lowering on few-bit activations.
In this work, we show how these problems can be addressed by absorbing floating point operations into thresholds (\emph{streamlining}), using a bit-serial formulation for handling few-bit integer matrix multiplications and channel-interleaved lowering.

\begin{table}
	\caption{Accuracy of a state-of-the-art QNN \cite{hwgq}.}
	\footnotesize
	\begin{tabular}{cccc}
		\toprule
		Dataset & Network & Floating Point & \qnn{1}{2} HWGQ \cite{hwgq} \\
		 &  & top-1 (top-5) & top-1 (top-5) \\
		\midrule
		ImageNet & AlexNet & 58.5\% (81.5\%) & 52.7\% (76.3\%) \\
		ImageNet & GoogLeNet & 71.4\% (90.5\%) & 63.0\% (84.9\%) \\
		ImageNet & VGG-like & 69.8\% (89.3\%) & 64.1\% (85.6\%) \\
		CIFAR-10 & VGG-like & 93.2\% & 92.5\% \\
		\bottomrule
	\end{tabular}
	\label{tab:hwgq-accuracy}
\end{table}

\section{Streamlined QNNs}

Even for layers with uniform-quantized input activations and weights, state-of-the-art QNN methods use some floating point computation in the forward pass to improve the accuracy.
Although these layers do not typically contain a large amount of computation, they may still incur slowdowns on devices where floating point operations are expensive and increase the memory footprint of the QNN by adding floating point parameters.
Three such examples from state-of-the-art QNN methods are:

\begin{enumerate}
	\item \textbf{Batch normalization.}
	Almost all state-of-the-art QNN techniques, including BinaryNet \cite{binarynet}, XNOR-Net \cite{xnornet} and HWGQ \cite{hwgq}, use batch normalization to obtain zero mean and unit variance prior to quantizing activations.
	The normalization parameters $\mu$ and $i$ are floating point values obtained during network training.

	\item \textbf{$\alpha$-scaling.}
	To better approximate the full-precision results using quantized operations, both XNOR-Net \cite{xnornet} and HWGQ \cite{hwgq} use $\alpha$-scaling.
	This involves multiplying the quantized matrix multiplication result with $\alpha$, which is a floating point vector containing the average L1-norm of each row of the weight matrix prior to quantization.

	\item \textbf{Non-integer quantization levels.}
	The chosen quantization levels in a QNN may be floating point values to best approximate the underlying value distribution.
	For instance, the state-of-the-art QNNs produced by HWGQ \cite{hwgq} use the following function for 2-bit uniform quantization:
	\[
	\mathrm{HWGQ}(x) = \left\{\begin{array}{lr}
	0, & \text{for } x \leq t_0\\
	0.538, & \text{for } 0 < x \leq 0.807\\
	1.076, & \text{for } 0.807 < x \leq 1.345\\
	1.614, & \text{for } 1.345 < x
	\end{array}\right\}
	\]
\end{enumerate}

\begin{figure}
\centering
\includegraphics[width=0.95\linewidth]{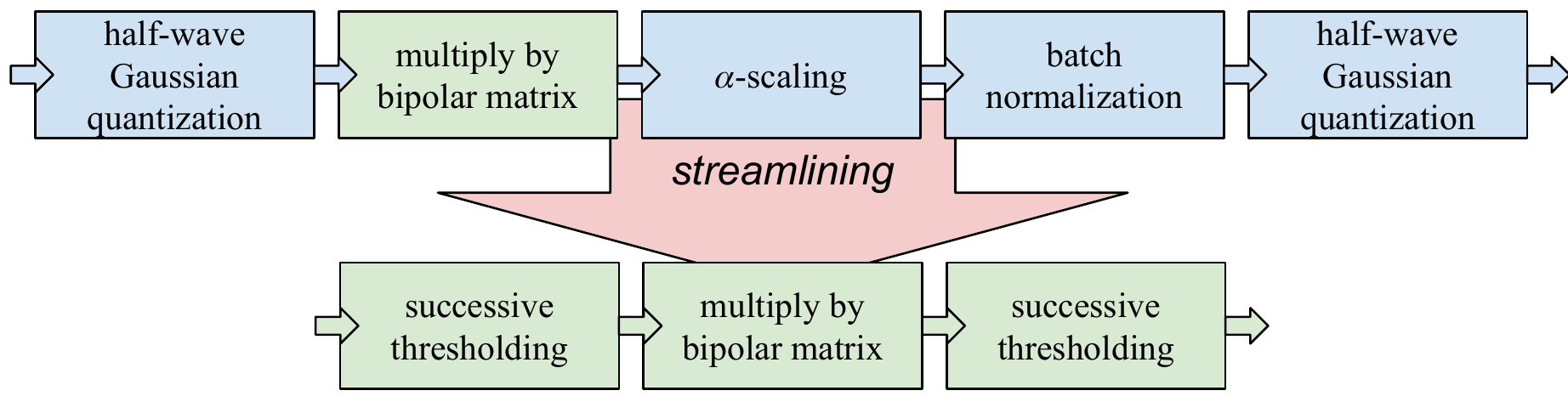}
\caption{Streamlining an HWGQ network. Blue and green color indicate floating point and integer data, respectively.}
\label{fig:hwgq-streamlining-overview}
\end{figure}

\subsection{The Streamlining Algorithm}
Through a process we call \emph{streamlining}, we show how the forward pass through any QNN layer with uniform-quantized activations and weights can be computed using only integer operations.
This consists of the following three steps:

\subsubsection{Quantization as successive thresholding} Given a set of threshold values $t = \{t_0, t_1 \dots t_{n}\}$,
the successive thresholding function $T(x, t)$ maps any real number $x$ to an integer in the interval $[0, n]$, where the returned integer is the number of thresholds that $x$ is greater than or equal to:

\[
T(x, t) = \left\{\begin{array}{lr}
0, & \text{for } x \leq t_0\\
1, & \text{for } t_0 < x \leq t_1\\
\dots & \dots \\
n-1, & \text{for } t_{n-2} < x \leq t_{n-1}\\
n, & \text{for } t_{n-1} < x
\end{array}\right\}
\]

Any uniform quantizer $Q(x)$ can be expressed as successive thresholding followed by a linear transformation such that $Q(x) = a \cdot T(x)+b$.
As an example, the 2-bit uniform HWGQ quantizer can be expressed as $\mathrm{HWGQ}(x) = 0.538 \cdot T(x, t)$ with $t_0 = 0, t_1 = 0.807, t_2 = 1.345$.
It should be noted that this technique is only economical for few-bit activations, since the number of thresholds grows exponentially with the activation bitwidth.

\subsubsection{Moving and collapsing linear transformations}
Any sequence of linear transformations can be collapsed into a single linear transformation.
We can first move all floating point linear operations to be positioned \emph{between} the quantized matrix operation and the activation quantization, then collapse them into a single linear transformation.
For the example in Figure \ref{fig:hwgq-streamlining-overview}, the linear transformation $ax+b$ for the previous layer's activation quantization can be moved past the bipolar matrix multiplication, since $W\cdot (ax+b) = a\cdot(Wx) + Wb$, forming a sequence together with the $\alpha$-scaling and batch normalization.
Afterwards, this sequence of three linear transformations can be reduced to a single linear transformation.

\subsubsection{Absorbing linear operations into thresholds}
The final step in the streamlining process is to update the threshold values as $t_i \gets (t_i - b) / a$ using the parameters $a, b$ of the linear transformation.
Observe that in the inequality $t_0 < x \leq t_1$ we can substitute $ax+b$ as the variable, and rewrite it as $(t_0 - b) / a < x \leq (t_1 - b) / a$.
By updating each threshold in this manner, we can remove the floating point linear transformations completely and feed the result of the quantized matrix operation directly into the successive thresholding layer.
Furthermore, if the input to the quantized matrix operation is known to be integer (i.e. the previous layer's activations were also quantized), each threshold can be simply rounded up to the nearest integer without changing the produced results.

\section{Inference with Few-Bit Weights and Activations on Mobile CPUs}
The dominating computation in QNN inference is convolutions between feature maps and kernels expressed as few-bit integers, which can be \emph{lowered} \cite{convlowering} to matrix-matrix multiplication between few-bit integer matrices.
Both the lowering and the matrix multiplications can be carried out by casting all operands to 8-bit integers, which are natively supported by most ISAs today.
Libraries such as Google's gemmlowp \cite{gemmlowp}, which has been used to deploy DNNs on mobile devices, offer high-performance 8-bit matrix multiplications.
However, using 8-bit operands to carry out few-bit integer operations can be wasteful.
For instance, using 8-bit operations to compute a \qnn{2}{2} matrix product would insert six zero bits into each operand, thus unnecessarily increasing the memory footprint by $4\times$.
Here, we provide alternatives that take advantage of few-bit integers for both the lowering and matrix multiplication operations.

\subsection{Few-Bit Integer Matrix Multiplication}
\label{sec:fewbitgemm}
To perform efficient few-bit integer matrix multiplication, we propose to use commonly-supported bitwise operations in \emph{bit-serial} fashion.
We will first describe how this is done for the \qnn{1}{1} case, then generalize the method to \qnn{w}{a}.

\label{sec:bingemm}
\textbf{The \qnn{1}{1} case.}
Binary matrix multiplication, which we refer to as \textsc{BinaryGEMM}, can be used for the case where each weight and activations can be represented using a single bit.
Previous work \cite{fptbnn, binarynet, xnornet} discussed how binary dot products can be implemented using bitwise XNOR followed by popcount (counting the number of set bits) operations.
Most modern processors provide an instruction for popcount, which enables fast \textsc{BinaryGEMM} implementations even on mobile CPUs.
Note that very high performance (in the trillion-operations per second range) on these operations can also be achieved with FPGAs \cite{finn,fptbnn} and GPGPUs \cite{binarynet}.
Although XNOR-popcount is only applicable for matrices with $\{-1, +1\}$ binary elements, it is possible to extend this idea to $\{0, 1\}$ binary elements by using bitwise AND instead of XNOR as shown in Algorithm \ref{alg:bingemm}.

\begin{algorithm}[t]
	\centering
	\footnotesize
	\begin{algorithmic}
		\algrenewcommand\algorithmicindent{1.0em}%
		\Function{BinaryGEMM}{$W, A, \mathrm{res}, \alpha$}
		\For{$r \gets 0 \dots rows-1$}
			\For{$c \gets 0 \dots cols-1$}
				\For{$d \gets 0 \dots \myceil{depth/wordsize}-1$}
					\State $res[r][c] \pluseq \alpha \cdot \Call{Popcount}{W(r, d) \mathrel{\&} A(c, d)} $
				\EndFor
			\EndFor
		\EndFor

		\EndFunction
	\end{algorithmic}
	\caption{\qnn{1}{1} GEMM using AND-popcount.}
	\label{alg:bingemm}
\end{algorithm}

\textbf{The \qnn{w}{a} case.}
\label{sec:bsgemm}
We now leverage \textsc{BinaryGEMM} as a building block for implementing few-bit integer matrix multiplication.
We can rewrite the $w$-bit matrix $W$ as a weighted sum $\sum_{i=0}^{w-1} 2^i \cdot$\bitplane{W}{i}, where \bitplane{W}{i} is the binary matrix formed by taking bit $i$ of each element of $W$, also referred to as a \emph{bit plane}.
In this manner, the product of two few-bit integer matrices can be written as a weighted sum of the pairwise products of their bit planes, i.e. $W \cdot A = \sum_{i=0}^{w-1} \sum_{j=0}^{a-1} 2^{i+j} \cdot$\bitplane{W}{i}$\cdot$\bitplane{A}{j}.

Algorithm \ref{alg:bsgemm} uses this observation to formulate few-bit integer matrix multiplication between the $w$-bit weight matrix $W$ and the $a$-bit activation matrix $A$.
This is a \emph{vectorized bit-serial} operation, since the contributions to each result element are computed between two bit positions at a time, but the bitwise operations inside \textsc{BinaryGEMM} operate on vectors of bits.
In this manner, we are able to take advantage of the full width of the processor datapath without introducing a large number of zero bits inside operations regardless of the values of $w$ and $a$.
The time taken by \ours{} will be proportional to $w \cdot a$, with \qnn{1}{1} executing fastest.

\begin{algorithm}[t]
	\centering
	\footnotesize
	\begin{algorithmic}
		\algrenewcommand\algorithmicindent{1.0em}%
		\Function{BitSerialGEMM}{$W, A, \mathrm{res}$}
		\For{$i \gets 0 \dots w-1$}
		\For{$j \gets 0 \dots a-1$}
		\State $\mathrm{sgnW} \gets (i == w-1 \mathrel{?} -1 : 1)$
		\State $\mathrm{sgnA} \gets (j == a-1 \mathrel{?} -1 : 1)$
		\State \Call{BinaryGEMM}{\bitplane{W}{i}, \bitplane{A}{j}, res, $\mathrm{sgnW} \cdot \mathrm{sgnA} \cdot 2^{i+j}$}
		\EndFor
		\EndFor
		\EndFunction
	\end{algorithmic}
	\caption{Signed \qnn{w}{a} GEMM using \textsc{BinaryGEMM}.}
	\label{alg:bsgemm}
\end{algorithm}

\subsection{Lowering with Few-Bit Activations}
Lowering convolutions \cite{convlowering} allows taking advantage of optimized matrix-matrix multiplications, which is the approach we take in this work.
Memory usage is typically a concern for lowering due to duplicated pixels.
In theory, QNNs do not suffer as much from this problem since quantized activations use much fewer bits per pixel, but taking advantage of this on a CPU can be tricky.
Namely, the lowering process itself (often called \emph{im2col}) requires accessing the feature map data in a "sliding window" fashion, which may require bit masking and shifting operations that decrease performance.
Representing each few-bit activation as an 8-bit value avoids this problem, but introduces many unused zero padding bits.
\begin{figure}
	\includegraphics[width=0.8\linewidth]{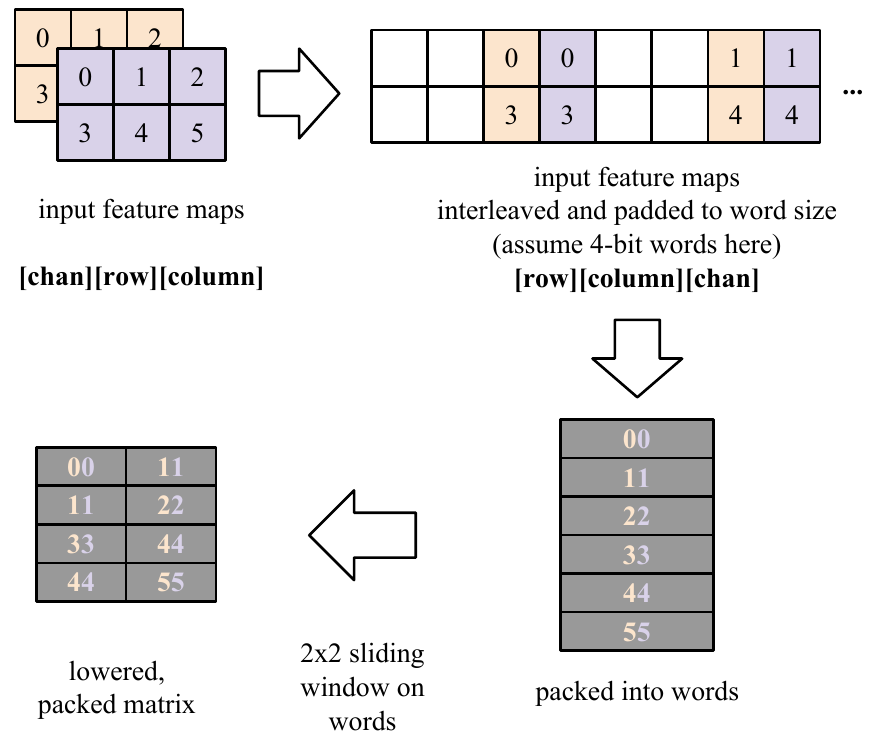}
	\caption{Interleaved lowering.}
	\label{fig:interleaved}
\end{figure}

We propose an alternative, which we refer to as \emph{interleaved lowering}, that uses a bit-serial, channel-interleaved data layout as illustrated in Figure \ref{fig:interleaved}.
Each pixel, which may span one or more CPU words, contains the bits from one bit position across activations from all channels, padded to the nearest word boundary.
Afterwards, the lowering can be performed on the granularity of entire CPU words, with one im2col per activation bit.

\section{Evaluation}
We implemented \ours{} using ARM NEON intrinsics in C++, with register blocking and L1 cache blocking to achieve higher performance.
We compare against the gemmlowp library \cite{gemmlowp}, which utilizes hand-optimized inline assembly for 8-bit matrix multiplications.
All reported results are measured on a single ARM Cortex-A57 core running at 1.9~GHz on the Nvidia Jetson~TX1 board.
We use 8-bit native matrix multiplications provided by gemmlowp \cite{gemmlowp} as the baseline alternative to \ours{}.

\subsection{Matrix Multiplication Microbenchmarks}
As matrix multiplication accounts for the majority of time in neural network inference, we start by evaluating \ours{} on matrix multiplication microbenchmarks.
For a (rows, depth, cols) operation that takes $T$~nanoseconds, we report the performance in integer giga-operations per second (GOPS) measurement as $(2\cdot\mathrm{rows}\cdot\mathrm{depth}\cdot\mathrm{cols}) \mathrel{/} T$ by averaging over a runtime of 10~s.

\subsubsection{Compute-Bound Performance}
\label{sec:peakperf}
To measure the maximum achievable performance with our implementation, we use the largest matrices that still fit into the L1 cache.
For gemmlowp, we observed a peak performance of 22~GOPS.
For \ours{} on \qnn{1}{1} (binary matrices), we observed a peak performance of 150~GOPS, which is $6.8\times$ faster than using 8-bit operands.
As expected, the performance linearly decreases with more bits of precision: 77~GOPS for \qnn{1}{2}, 50~GOPS for \qnn{1}{3}, 34~GOPS for \qnn{2}{2} and 23~GOPS for \qnn{2}{3}.
Thus, for this particular platform, \ours{} is faster than using 8-bit operations for \qnn{w}{a} with $w \cdot a \leq 6.8$ when working with in-cache matrices.

\subsubsection{Performance vs Matrix Size}
\label{sec:perfvssize}
\begin{figure}
	\centering
	\footnotesize
	\input{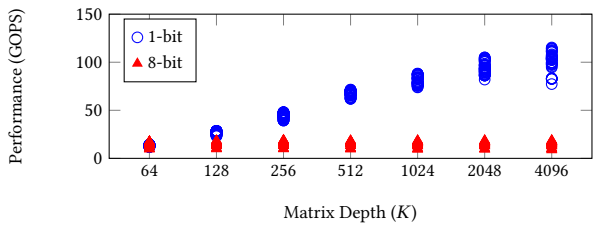}
	\caption{Log-linear plot of performance versus depth ($K$).}
	\label{fig:depthvsgops}
\end{figure}

To investigate how performance is influenced by the dimensions of a $M \times N \times K$ matrix multiplication, we performed a sweep of different sizes between $2^6$ and $2^{12}$ in each dimension using both gemmlowp and \ours{} with \qnn{1}{1}.
Figure \ref{fig:depthvsgops} presents a scatter plot of the performance with increasing depth ($K$).
We observe that \ours{} performance is sensitive to the depth ($K$) dimension.
For small matrix sizes, there is little or no performance advantage over gemmlowp, which should be taken into consideration when choosing the execution method for each layer.
\ours{} quickly becomes faster with increasing depth and becomes advantageous over gemmlowp, up to $6.6\times$ faster than gemmlowp for a $64 \times 1024 \times 4096$ multiplication.
With $K \geq 2048$, we observe decreased performance for larger $M$ and $N$ values in \ours{} due to increased cache misses, which can be addressed by adding more levels of blocking to the implementation.

\subsection{Quantized AlexNet}
To assess the benefits of the techniques discussed for QNN deployment, we developed a version of Caffe with support for quantized layers.
Each quantized layer can be configured individually to use either gemmlowp or \ours{} as the execution engine.
We use a quantized \qnn AlexNet from \cite{hwgq} with batch size 1 as a benchmark, with \qnn{8}{8} for the first layer, \qnn{8}{2} for the last layer, and \qnn{1}{2} for all other matrix layers.
The first and last layer are computed using gemmlowp in 8-bit precision to preserve accuracy.
For non-matrix layers such as thresholding and max pooling which constitute a tiny portion of the total compute, we use regular floating point operations.
We evaluated the performance for the following combinations of techniques:

\begin{itemize}
	\item \textbf{baseline:} No streamlining, all layers using gemmlowp.
	\item \textbf{bsgemm:} Streamlining, all but the first and last layer using \ours{}, first and last layer in gemmlowp.
	\item \textbf{bsgemm-intl:} Streamlining and interleaving, all but the first and last layer using \ours{} .
\end{itemize}

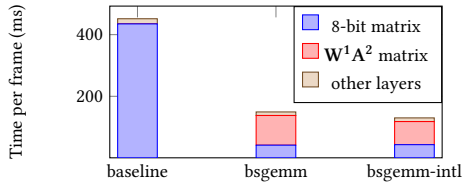
\begin{figure}
	\footnotesize
	\begin{tikzpicture}
\begin{axis}[
ybar stacked,
bar width=15pt,
enlargelimits=0.1,
legend style={at={(1,1)},anchor=north east},
width=6cm, height=3.6cm,
ylabel={Time per frame (ms)},
symbolic x coords={baseline, bsgemm, bsgemm-intl},
xtick=data
]

\addplot+[ybar] plot coordinates {
	(baseline, 434.8346)
	(bsgemm, 41.4811)
	(bsgemm-intl, 43.105)
};

\addplot+[ybar] plot coordinates {
	(baseline, 0)
	(bsgemm, 96.27407)
	(bsgemm-intl, 74.99308)
};	

\addplot+[ybar] plot coordinates {
		(baseline, 16.7594)
		(bsgemm, 11.52183)
		(bsgemm-intl, 11.86192)
};		

\legend{\strut 8-bit matrix, \strut \qnn{1}{2} matrix, \strut other layers}
\end{axis}
\end{tikzpicture}
	\caption{Overall performance for quantized AlexNet.}
	\label{fig:breakdown}
\end{figure}

\subsubsection{Overall performance}
Figure \ref{fig:breakdown} compares the time per frame with the three techniques and presents a basic breakdown of time cost.
Overall, bsgemm achieves a $3\times$ speedup over the baseline, and bsgemm-intl further improves this to $3.5\times$.
Speedups from bsgemm are limited by the presence of 8-bit first/last layers, which account for 33\% of the execution time in bsgemm-intl.
Also quantizing those layers further would bring further performance benefits.
The current throughput is 2.2, 6.7 and 7.7 frames per second respectively for \textbf{baseline}, \textbf{bsgemm} and \textbf{bsgemm-intl}, and the performance can be further improved by multi-core parallelism and code optimization.

\begin{table}
	\centering
	\caption{Time cost breakdown for \qnn{1}{2} AlexNet with batch size 1. Best numbers for each row are highlighted.}
	\label{tab:detailedbreakdown}
	\footnotesize
	\begin{tabular}{@{}cccccc@{}}
		\toprule
&		Operation & \multicolumn{3}{c}{Time (ms)} & Speedup \\
		\cmidrule(r){3-5}
&		   & baseline       & bsgemm          & bsgemm-intl             &          \\
		\midrule
		\parbox[t]{2mm}{\multirow{10}{*}{\rotatebox[origin=c]{90}{convolutional}}} &
		lowering & \textbf{6.7} & \textbf{6.7} & \textbf{6.7} & 1$\times$ \\
&		(96, 363, 3025) & \textbf{20.7} & \textbf{20.7} & \textbf{20.7} & 1$\times$ \\
&		lowering & 8.7 & 15 & \textbf{0.9} & 10$\times$ \\
&		(256, 2400, 729) & 90.3 & 23.7 & \textbf{23.7} & 3$\times$ \\
&		lowering & 2.4 & 3.7 & \textbf{0.2} & 12$\times$ \\
&		(384, 2304, 169) & 32.2 & 8.3 & \textbf{8.3} & 4$\times$ \\
&		lowering & 3.5 & 5.7 & \textbf{0.3} & 10$\times$ \\
&		(384, 3456, 169) & 48 & 10.7 & \textbf{10.7} & 5$\times$ \\
&		lowering & 3.5 & 5.7 & \textbf{0.3} & 10$\times$ \\
&		(256, 3456, 169) & 35.8 & 7.3 & \textbf{7.3} & 5$\times$ \\
		\midrule
		\parbox[t]{2mm}{\multirow{3}{*}{\rotatebox[origin=c]{90}{FC}}}
&		(4096, 9216, 1)  & 114.7 & 2.3 & \textbf{2.3} & 50$\times$ \\
&		(4096, 4096, 1)  & 52.9 & 1.1 & \textbf{1.1} & 50$\times$ \\
&		(1000, 4096, 1)  & \textbf{13} & \textbf{13} & \textbf{13} & 1$\times$	\\
		\bottomrule

	\end{tabular}
\end{table}

\subsubsection{Detailed comparison.}
Table \ref{tab:detailedbreakdown} presents a detailed breakdown of time spent in lowering and matrix multiplication operations across AlexNet convolutional and fully-connected layers with different optimizations.
All matrix multiplications are indicated in parantheses as (rows, columns, depth), with the midline separating convolutional and fully-connected layers.
The advantage of using quantized operations for fully-connected layers is especially prominent, with speedups of up to $50\times$ owing to increased arithmetic intensity in matrix-vector multiplications.
For matrix-matrix multiplications in convolutional layers, the advantage of using \qnn{1}{2} \ours{} is around $4\times$.
When the matrix multiplies become faster, the overhead of lowering and bit packing costs become substantial, almost as much as the matrix-matrix multiplication time for earlier layers with large filters.
Fortunately, this can be remedied by interleaving, as indicated by the results in the \textbf{bsgemm-intl} column.
By taking advantage of packing bits prior to lowering, interleaved lowering can offer a $10\times$ speedup over the baseline.

\section{Conclusion and Future Work}
We have presented methods for efficient processing of QNNs on mobile CPUs via absorbing scaling factors into thresholds, channel-interleaved lowering, and bit-serial matrix multiplication.
Our results indicate that these methods can take better advantage of few-bit operations in QNNs, offering significant speedups over native 8-bit operations on mobile CPUs. As future work, we note that this approach can enable approximate computing by only considering the contributions from higher-order bits and taking advantage of bit-level sparsity.
Another use for this technique would be on-device training DNNs using low-precision gradients \cite{lowprecsgd}, which also requires low-precision matrix operations.
Our bit-serial matrix multiplication library is available on GitHub at https://git.io/vhshn.

\bibliographystyle{ACM-Reference-Format}
\bibliography{references}

\end{document}